# A Comparison of SVM against Pre-trained Language Models (PLMs) for Text Classification Tasks

Yasmen Wahba [1](✉), Nazim Madhavji[1], and John Steinbacher [2]

[1] Western University, London ON, Canada
ywahba2@uwo.ca, nmadhavji@uwo.ca
[2] IBM Canada, Toronto, ON, Canada
jstein@ca.ibm.com

**Abstract.** The emergence of pre-trained language models (PLMs) has shown great success in many Natural Language Processing (NLP) tasks including text classification. Due to the minimal to no feature engineering required when using these models, PLMs are becoming the de facto choice for any NLP task. However, for domain-specific corpora (e.g., financial, legal, and industrial), fine-tuning a pre-trained model for a specific task has shown to provide a performance improvement. In this paper, we compare the performance of four different PLMs on three public domain-free datasets and a real-world dataset containing domain-specific words, against a simple SVM linear classifier with TFIDF vectorized text. The experimental results on the four datasets show that using PLMs, even fine-tuned, do not provide significant gain over the linear SVM classifier. Hence, we recommend that for text classification tasks, traditional SVM along with careful feature engineering can provide a cheaper and superior performance than PLMs.

**Keywords:** Text Classification, Pre-trained Language Models, Machine Learning, Domain-specific Datasets, Natural Language Processing

## 1   Introduction

Text classification is the task of classifying text (e.g., tweets, news, and customer reviews) into different categories (i.e., tags). It is a challenging task especially when the text is 'technical'. We define 'technical' text in terms of the vocabulary used to describe a given document, e.g., classifying health records, human genomics, IT discussion forums, etc. These kinds of documents require special pre-processing since the basic NLP pre-processing steps may remove critical words necessary for correct classification, resulting in a performance drop of the deployed system [1].

Recently, pre-trained language models (PLMs) such as BERT [2] and ELMO [3] have shown promising results in several NLP tasks, including spam filtering, sentiment analysis, and question answering. In comparison to traditional models, PLMs require less feature engineering and minimal effort in data cleaning. Thus becoming the consensus for many NLP tasks [4].

With an enormous number of trainable parameters, these PLMs can encode a substantial amount of linguistic knowledge that is beneficial to contextual representations [4]. For example, word polysemy (i.e., the coexistence of multiple meanings for a word or a phrase –e.g., 'bank' could mean 'river bank' or 'financial bank') in a domain-free text.

In contrast, in a domain-specific text that contains technical jargon, a word has a more precise meaning (i.e., monosemy) [5]. For example, the word 'run' in an IT text would generally only mean 'execute' and not 'rush'. Thus, it appears that domain-specific text classification will likely not benefit from the rich linguistic knowledge encoded in PLMs.

Despite the widespread use of PLMs in a broad range of downstream tasks, their performance is still being evaluated by researchers for their drawbacks [6]. For example: (i) the large gap between the pre-training objectives (e.g., predict target words) and the downstream objectives (e.g., classification) limits the ability to fully utilize the knowledge encoded in PLMs [7], (ii) the high computational cost and the large set of trainable parameters make these models impractical for training from scratch, (iii) dealing with rare words is a challenge for PLMs [8], and (iv) the performance of PLMs may not be generalizable [9].

Thus, this paper evaluates the performance of different pre-trained language models (PLMs) against a linear Support Vector Machine (SVM) classifier. The motivation for this comparative study is rooted in the fact that: (i) while PLMs are being used in text classification tasks [10][11], they are more computationally expensive than the simpler SVMs, and (ii) PLMs have been used predominantly on public or domain-free datasets and it is not clear how they fare against simpler SVMs on domain-specific datasets.

The findings of our study suggest that the problem of classifying domain-specific or generic text can be addressed efficiently using old traditional classifiers such as SVM and a vectorization technique such as TFIDF bag-of-words that do not involve the complexity found in neural network models such as PLMs. To the best of our knowledge, no such comparative analysis has so far been described in the scientific literature.

The rest of the paper is organized as follows. Section 2 describes related work. Section 3 describes the empirical study. Section 4 presents the research results. Section 5 concludes the paper.

## 2   Related Work

In this section, we give an overview of the existing literature on the applications of PLMs and some of the drawbacks reported.

Pre-trained language models (PLMs) are deep neural networks trained on unlabeled large-scale corpora. The motivation behind these models is to capture rich linguistic knowledge that could be further transferred to target tasks with limited training samples (i.e., fine-tuning). BERT [2], XLM [12], RoBERTa [13], and XLNet [14] are examples of PLMs that have achieved significant improvements on a large number of NLP tasks (e.g., question answering, sentiment analysis, text generation).

Nevertheless, the performance of these models on domain-specific tasks was questioned [15] as these models are trained on general domain corpora such as Wikipedia, news websites, and books. Hence, fine-tuning or fully re-training PLMs for downstream tasks has become a consensus. Beltagi et al. [16] released SciBERT that is fully retrained on scientific text (i.e., papers). Lee et al. [17] released BioBERT for biological text. Similarly, Clinical BERT [18][19] was released for clinical text, and FinBERT [20] for the financial domain.

Other researchers applied PLMs by fine-tuning the final layers to the downstream task. For example, Elwany et al. [21] report valuable improvements on legal corpora after fine-tuning. Lu [22] fine-tuned RoBERTa for Commonsense Reasoning and Tang et al. [23] fine-tuned BERT for multi-label sentiment analysis in code-switching text. Finally, Yuan et al. [24] fine-tuned BERT and ERNIE [25] for the detection of Alzheimer's Disease.

However, Gururangan et al. [15] show that simple fine-tuning of PLMs is not always sufficient for domain-specific applications. Their work suggests that the second phase of pre-training can provide significant gains in task performance. Similarly, Kao et al. [26] suggest that duplicating some layers in BERT prior to fine-tuning can lead to better performance on downstream tasks.

Another body of research focuses on understanding the weaknesses of PLMs by either applying them to more challenging datasets or by investigating their underlying mechanisms. For example, McCoy et al. [9] report the failure of BERT when evaluated on the HANS dataset. Their work suggests that evaluation sets should be drawn from a different distribution than the train set. Also, Schick et al. [8] introduce WNLaMPro (WordNet Language ModelProbing) dataset to assess the ability of PLMs to understand rare words. Lastly, Olga et al. [27] show redundancy in the information encoded by different heads in BERT, and manually disabling attention in certain heads will lead to performance improvement.

This paper adds to the growing literature on evaluating PLMs. In particular, our investigative question is: How does a linear classifier such as SVM compare against the state-of-the-art PLMs on both general and technical domains?

## 3 Empirical Study

In this section, we describe the empirical study that we conducted. In particular, we describe the infrastructure used, the datasets, and the different PLMs used. Finally, we describe the SVM algorithm used, and the pre-processing steps done prior to applying SVM. The experimental algorithms are written in Python 3.8.3. The testing machine is Windows 10 with an Intel Core i7 CPU 2.71 GHz and 32GB of RAM.

### 3.1 Text Classification Datasets

Our experiments were evaluated on four datasets:
1. BBC News [28]: a public dataset originating from BBC News. It consists of 2,225 documents, categorized into 5 groups, namely: business, entertainment, politics, sport, and tech.
2. 20NewsGroup [29]: a public dataset consisting of 18,846 documents, categorized into 20 groups.

43. Consumer Complaints [30]: a public benchmark dataset published by the Consumer Financial Protection Bureau; it is a collection of complaints about consumer financial products and services. It consists of 570,279 documents categorized into 15 classes.

4. IT Support tickets: a private dataset obtained from a large industrial partner. It is composed of real customer issues related to a cloud-based system. It consists of 194,488 documents categorized into 12 classes.

Table 1 summarizes the properties of the four datasets.

**Table 1.** Dataset properties

| Dataset | # of classes | # of instances | # of features (n-gram=1) | # of features (n-gram=3) |
| --- | --- | --- | --- | --- |
| BBC News | 5 | 2,225 | 26,781 | 811,112 |
| 20NewsGroup | 20 | 18,846 | 83,667 | 2,011,358 |
| Consumer Complaints | 15 | 570,279 | 53,429 | 6,112,905 |
| IT Support tickets | 12 | 194,488 | 16,011 | 3,185,796 |

The IT Support tickets dataset will be referred to hereon as the 'domain-specific' dataset. This dataset suffers from a severe imbalance as seen in Fig. 1. However, we prefer to avoid the drawbacks of sampling techniques [31][32] and keep the distribution as is.

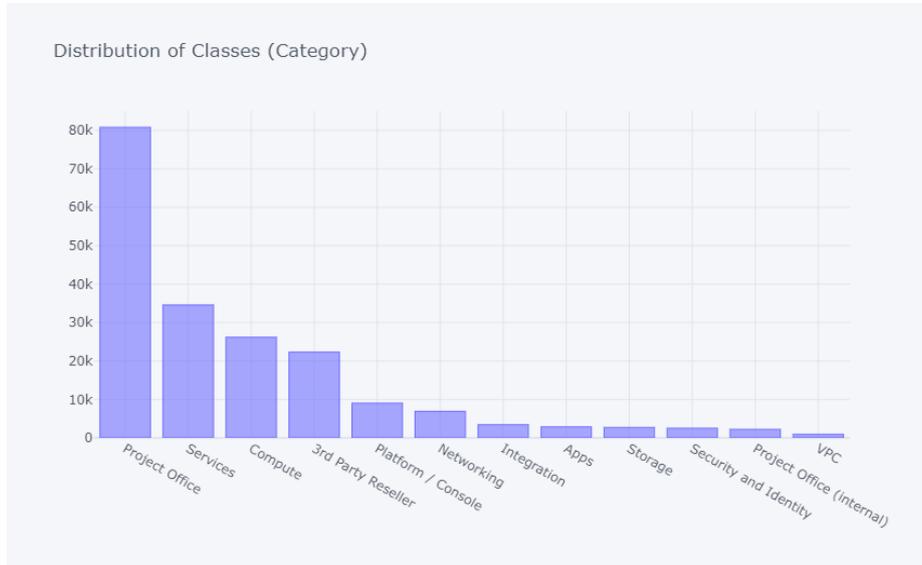

**Fig. 1.** Class distribution of the domain-specific dataset showing imbalance

Another problem with this dataset is the presence of a large number of technical words (i.e., jargon) related to the Cloud terminologies (e.g., Bluemix, Kubernetes, Iaas, Vmware, etc.). These words are not found in the PLMs vocabulary and hence, they get broken down into subwords using a subword tokenization algorithm. For instance, BERT uses a WordPiece tokenizer [33] which handles non-technical words quite well. However, we notice that it fails to tokenize technical words and domain-specific abbreviations in our domain-specific dataset. For example:

"Kubernetes" ⇒ ['ku', '##ber', '##net', '##es']
"configuration" ⇒ "config" ⇒ ['con', '##fi', '##g']

### 3.2 Pre-trained Language Models (PLMs)

The following PLMs were considered for this study:

1. BERT [2]: A widely used pre-training language model that is based on a bidirectional deep Transformer as the main structure. BERT achieved state-of-the-art results on 11 different NLP tasks including question answering and named entity recognition (NER).
2. DistilBERT [34]: A lighter, smaller, and faster version of BERT. By reducing the size of the BERT model by 40%, while keeping 97% of its language understanding capability, it's considered 60% faster than BERT.
3. RoBERTa [13]: One of the successful variants of BERT that achieved impressive results on many NLP tasks. By changing the MASK pattern, discarding the NSP task, and using a larger batch size and longer training sentences.
4. XLM [12]: Designed specifically for cross-lingual classification tasks by leveraging bilingual sentence pairs. XLM uses a known pre-processing technique (BPE) and a dual-language training mechanism.

For this study, we fine-tuned all the PLMs to the domain-specific dataset and the three generic datasets. In all our experiments, we use the following hyperparameters for fine-tuning: maximum sequence length of 256, adam learning rate (lr) of $1e^{-5}$, batch size of 16, and a train-test split ratio of 80:20.

**Support Vector Machines (SVM).** A Support Vector Machine is a popular supervised margin classifier, reported as one of the best algorithms for text classification [35] [36]. We chose the LinearSVC algorithm in the *Scikit-learn* library [37], which implements a one-versus-all (OVA) multi-class strategy. This algorithm is suitable for high-dimensional datasets and is characterized by a low running time [38].

Unlike PLMs, traditional machine learning models require pre-processing data cleaning steps. In our study, we used the following pre-processing steps on the four datasets: (i) removing missing data; (ii) removing numbers and special characters; (iii) lower casing; (iv) tokenization; (v) lemmatization; and (vi) word vectorization using TFIDF[1].

It is important to note that when applying the TFIDF vectorizer, we tried different N-grams. An 'N-gram' is simply a sequence of N words that predicts the occurrence of a word based on the occurrence of its (N – 1) previous words. The default setting is Unigrams. In our study, we used trigrams which means that we included feature vectors consisting of all unigrams, bigrams, and trigrams.

## 4 Results

In this section, we discuss the results of applying four different fine-tuned PLMs (i.e., BERT, DistilBERT, RoBERTa, XLM) and a linear SVM classifier on the four datasets described in Section 3.1.

Table 2 shows the F1-scores obtained when applying the four PLMs and a linear SVM classifier on the four datasets. When evaluating PLMs, we used 3 epochs because we observed that when the number of epochs exceeds 3, the training loss decreases with each epoch and the validation loss increases. This translates to overfitting. Thus, all our experiments are run for 3 epochs only.

For the domain-specific dataset, it is clear how the linear SVM achieves a comparable performance (0.79) as any of the fine-tuned PLMs. Similarly, for the BBC dataset, SVM surprisingly achieves the same F1-score (0.98) as RoBERTa on the third epoch. However, we expected that PLMs would significantly outperform SVM on general domain datasets.

For the 20NewsGroup, SVM outperformed all PLMs with an F1-score of 0.93. This accuracy score was a result of considering the meta-data (i.e., headers, footers, and quotes) as part of the text that is fed to the classifier. However, when we ignored the meta-data, there was a performance drop of 15%.

The last dataset is the Consumer Complaints which is the largest dataset (570,279 instances) as described in Table 1. The accuracy of the linear SVM (0.82) was very close to the highest accuracy of 0.85 obtained by BERT and RoBERTa. While 0.82 is very competitive, we believe there is room for improvement if feature selection techniques were considered as this dataset is characterized by a large feature set.

The accuracy scores of PLMs are generally higher on generic datasets that do not contain domain-specific or rare words. Also, we notice a small gap between the accuracy scores of all PLMs in the third epoch for all datasets.

In summary, the key points are:
- Linear SVM proved to be comparable to PLMs for text classification tasks.
- PLMs accuracy scores are generally higher on generic datasets.

---

[1] TFIDF stands for Term Frequency-Inverse Document Frequency, which is a combination of two metrics:

1. Term frequency (*tf*): a measure of how frequently a term *t*, appears in a document *d*.
2. Inverse document frequency(idf): a measure of how important a term is. It is computed by dividing the total number of documents in our corpus by the document frequency for each term and then applying logarithmic scaling on the result.

- The importance of feature engineering for text classification is highlighted by including meta-data.

**Table 2.** Comparison of four PLMs against SVM Linear classifier in terms of accuracy (F1-score)

| Dataset | Model | Epoch 1 | Epoch 2 | Epoch 3 |
|---|---|---|---|---|
| | | Accuracy (F1-score) | | |
| IT Support Tickets | *BERT* | 0.78 | 0.79 | 0.79 |
| | *DistilBERT* | 0.77 | 0.78 | 0.79 |
| | *XLM* | 0.77 | 0.79 | 0.79 |
| | *RoBERTa* | 0.77 | 0.78 | 0.79 |
| | *LinearSVM(n-gram=3)* | 0.79 | | |
| BBC | *BERT* | 0.97 | 0.97 | 0.97 |
| | *DistilBERT* | 0.97 | 0.97 | 0.97 |
| | *XLM* | 0.88 | 0.96 | 0.97 |
| | *RoBERTa* | 0.97 | 0.97 | 0.98 |
| | *LinearSVM(n-gram=3)* | 0.98 | | |
| 20News-Group | *BERT* | 0.85 | 0.91 | 0.92 |
| | *DistilBERT* | 0.82 | 0.90 | 0.90 |
| | *XLM* | 0.89 | 0.91 | 0.92 |
| | *RoBERTa* | 0.84 | 0.87 | 0.90 |
| | *LinearSVM* | 0.93 | | |
| Consumer Complaints | *BERT* | 0.83 | 0.84 | 0.85 |
| | *DistilBERT* | 0.82 | 0.84 | 0.84 |
| | *XLM* | 0.80 | 0.82 | 0.83 |
| | *RoBERTa* | 0.83 | 0.84 | 0.85 |
| | *LinearSVM* | 0.82 | | |

## 5 Conclusions

The study described in this paper compares the performance of several fine-tuned PLMs (see Section 3.2) against that of a linear SVM classifier (see Section 3.3) for the task of text classification. The datasets used in the study are: a domain-specific dataset of real-world support tickets from a large organization as well as three generic datasets (see Table 1).

To our surprise, we found that a pre-trained language model does not provide significant gains over the linear SVM classifier. We expected PLMs to outperform SVM on the generic datasets, however, our study indicates comparable performance for both models (see Table 2). Also, our study indicates that SVM outperforms PLMs on one of the generic datasets (i.e., 20NewsGroup).

Our finding goes against the trend of using PLMs on any NLP task. Thus, for text classification, we recommend prudence when deciding on the type of algorithms to use. Since our study seems to be the first comparative study of PLMs against SVM on generic datasets as well as on a domain-specific dataset, we encourage replication of this study to create a solid body of knowledge for confident decision-making on the choice of algorithms.